\documentclass[10pt,twocolumn,letterpaper]{article}

\usepackage{wacv}
\usepackage{times}
\usepackage{epsfig}
\usepackage{graphicx}
\usepackage{amsmath}
\usepackage{amssymb}
\usepackage{url}
\usepackage{color}
\usepackage{times}
\usepackage{amsmath}
\usepackage{booktabs}
\usepackage{multirow}
\usepackage{subcaption}

\makeatletter
\renewcommand*{\@fnsymbol}[1]{}
\makeatother

 \wacvfinalcopy 

\def\wacvPaperID{468} 

\ifwacvfinal\fi
\begin{document}

\title{Generating Handwritten Chinese Characters using CycleGAN}


\author{Bo Chang$^*$\thanks{* Authors contributed equally.} \hspace{2cm} Qiong Zhang$^*$ \hspace{2cm} Shenyi Pan \hspace{2cm} Lili Meng \\
University of British Columbia\\
{\tt\small \{bchang, qiong.zhang, shenyi.pan\}@stat.ubc.ca, menglili@cs.ubc.ca}
}

\maketitle
\ifwacvfinal\thispagestyle{empty}\fi

\begin{abstract}
Handwriting of Chinese has long been an important skill in East Asia. However, automatic generation of handwritten Chinese characters poses a great challenge due to the large number of characters. Various machine learning techniques have been used to recognize Chinese characters, but few works have studied the handwritten Chinese character generation problem, especially with unpaired training data. In this work, we formulate the Chinese handwritten character generation as a problem that learns a mapping from an existing printed font to a personalized handwritten style. We further propose DenseNet CycleGAN to generate Chinese handwritten characters. Our method is applied not only to commonly used Chinese characters but also to calligraphy work with aesthetic values. Furthermore, we propose content accuracy and style discrepancy as the evaluation metrics to assess the quality of the handwritten characters generated. We then use our proposed metrics to evaluate the generated characters from CASIA dataset as well as our newly introduced Lanting calligraphy dataset.
\end{abstract}


\section{Introduction}
\begin{figure}[ht]
    \begin{center}
    \includegraphics[width = 0.75\linewidth]{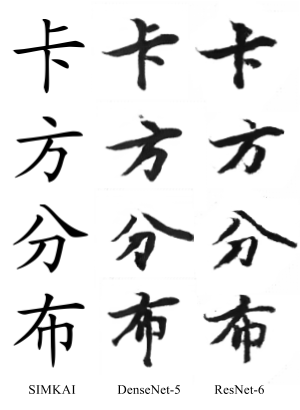}
    \end{center}
    \caption{\textbf{Generated Chinese characters in Wang Xizhi's style}. The characters (from top to bottom) in the first column is the source SIMKAI font, the second and the third columns are in the Wang Xizhi's style with each column generated from DenseNet-5 and ResNet-6 CycleGAN. These four characters are not in the Lanting calligraphy dataset. The four characters stand for $\chi^2$ distribution in Chinese.}
    \label{fig: wangxizhi}
\end{figure}

Chinese characters have been used continually over three millennia by more than a quarter of the world's population \cite{taylor2014writing}. The handwriting of Chinese character has long been one of the most fundamental skills in education, employment, communication, and everyday life in East Asia. For a long time, good handwriting or calligraphy has been considered not only as an artistic expression of language, but also as the supreme visual art as a means of self-expression and cultivation. 
As an example, Figure~\ref{fig: wangxizhi} shows some of our generated calligraphic characters.
These aesthetically pleasing calligraphic works usually needs years of dedication and practice.

In contrast to phonological languages that have very limited number of letters such as English, Chinese has more than $80,000$ logographic characters. Therefore, it is more challenging to design a personalized Chinese font than phonological languages. For example, only $26$ letters need to be designed for a personalized English font while for Chinese at least $3,000$ most commonly used characters need to be designed. To meet the demands of designing personalized Chinese font, methods that can automatically generate characters with personalized handwritten style based on a relatively small set of training characters are needed.

\begin{figure}
\[
\begin{array}{c|c}
\begin{array}{c}
x_i \quad \quad  \quad y_i\\
\left\{
\begin{array}{cc}
\includegraphics[width=0.07\linewidth]{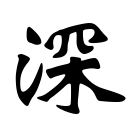}&\includegraphics[width=0.07\linewidth]{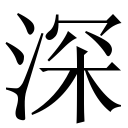}\\
\end{array}
\right\}\\
\left\{
\begin{array}{cc}
\includegraphics[width=0.07\linewidth]{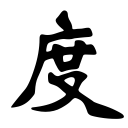}&\includegraphics[width=0.07\linewidth]{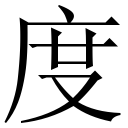}\\
\end{array}
\right\}\\
\left\{
\begin{array}{cc}
\includegraphics[width=0.07\linewidth]{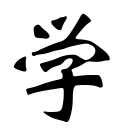}&\includegraphics[width=0.07\linewidth]{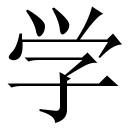}\\
\end{array}
\right\}\\
\left\{
\begin{array}{cc}
\includegraphics[width=0.07\linewidth]{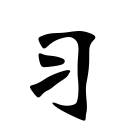}&\includegraphics[width=0.07\linewidth]{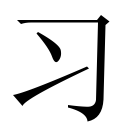}\\
\end{array}
\right\}\\
 \textbf{\vdots}
\end{array}     

& \begin{array}{cc}
    X & Y \\
\left\{
\begin{array}{c}
\includegraphics[width=0.08\linewidth]{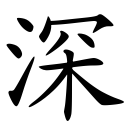} \\
 \includegraphics[width=0.08\linewidth]{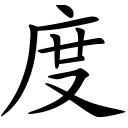}\\
 \includegraphics[width=0.08\linewidth]{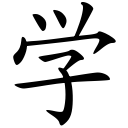}\\
 \includegraphics[width=0.08\linewidth]{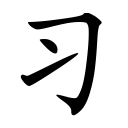}\\
 \textbf{\vdots}\\
\end{array}
\right\} & \left\{
\begin{array}{c}
\includegraphics[width=0.075\linewidth]{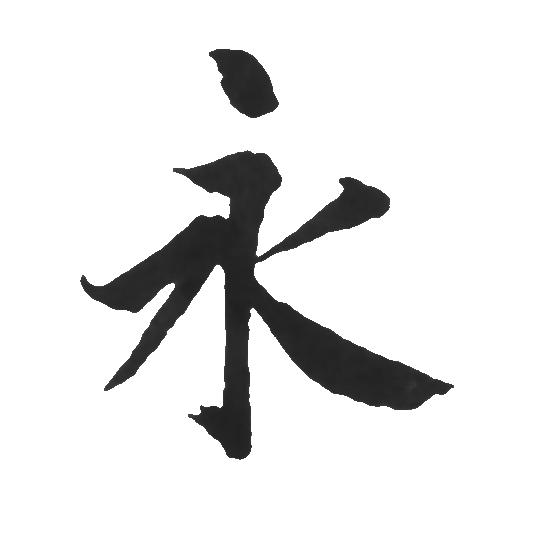} \\
 \includegraphics[width=0.075\linewidth]{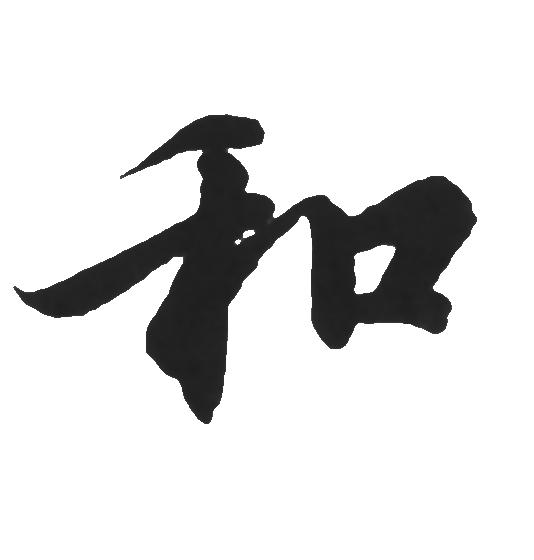}\\
 \includegraphics[width=0.075\linewidth]{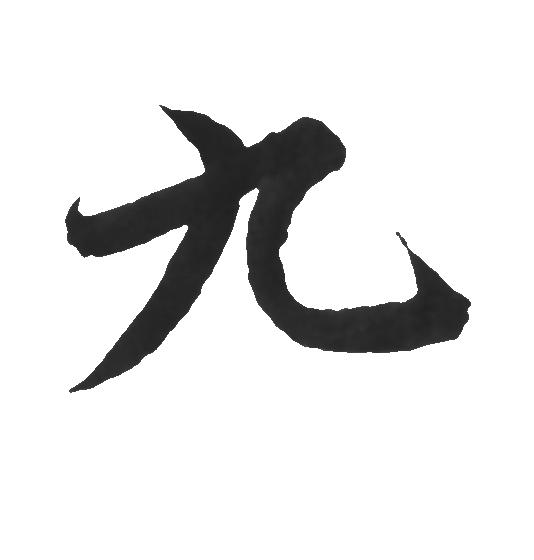}\\
 \includegraphics[width=0.075\linewidth]{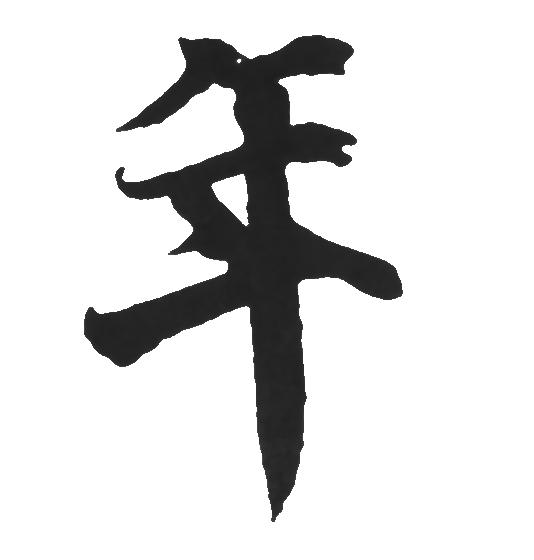}\\ 
 \textbf{\vdots}\\
\end{array}
\right\}
\end{array}\\
(a) \ \text{Paired}
&(b) \ \text{Unpaired}
\end{array}
\]
\caption{\textbf{Paired and unpaired training data.} (a) Paired training data consists of training examples
$\{x_i, y_i\}^N_{i=1}$, where there exists correspondence between $x_i$ and $y_i$. (b) We consider unpaired training data, where a source set $X$ and a target set {Y} exist, with no matching information for $x_i$ and $y_i$.}
\label{fig:pair_unpair}
\end{figure}

Although handwritten Chinese character generation is not as widely studied as character recognition, there are still approaches proposed for handwritten Chinese character generation. Most previous works rely on the hierarchical representation of simple strokes~\cite{xu2005automatic, xu2009automatic, liu2012automatic}. They decompose Chinese characters into strokes and then combine strokes to mimic the personalized writing style. As a result, such methods only focus on local representations of the characters rather than the overall style as a whole, and thus need to adjust the shapes, sizes, and positions of the strokes for every new character. In contrast, zi2zi~\cite{zi2zi} learns to transform fonts using pix2pix~\cite{isola2016image} with paired character images as the training data. However, in the task of generating handwritten Chinese characters, it is difficult to obtain a large set of paired training examples since it is infeasible to ask the user to write a large number of characters. 
The handwriting samples are also often isolated from a user's writing, without knowing the true labels of the characters.
Further, even the same character written by a user varies every time, which makes it more important to learn the overall style instead of mimicking every single character.
Therefore, it is more appropriate to use unpaired Chinese characters instead of paired data for the handwritten Chinese generation problem. Figure~\ref{fig:pair_unpair} provides an example of paired and unpaired training data. Paired training data contain images of the same character in both fonts, while unpaired data do not necessarily contain the same set of characters.

In this work, we formulate the Chinese handwritten character generation as a problem that learns a mapping from an existing printed font to a personalized handwritten style. We further propose a method based on unpaired image-to-image translation to solve this problem. 
Our main contributions are:
\begin{itemize} 
\item We propose to generate Chinese characters in a personalized handwritten style using DenseNet CycleGAN. 
\item We propose content accuracy and style discrepancy as the evaluation metrics to assess the quality of the generated characters.  
\item We demonstrate the efficacy of our proposed method on both CASIA dataset~\cite{liu2011casia} and our newly introduced Lanting calligraphy dataset. 
\end{itemize}

\section{Related work}
\subsection{Chinese character handwriting generation}
Chinese character generation has been studied since the beginning of the digital age~\cite{plamondon1998generation}. In the literature, Chinese character generation is mainly formulated as either an artistic calligraphy, typography generation problem~\cite{xu2005automatic, zi2zi,yang2016awesome, lian2012automatic}, or a personal handwriting generation problem \cite{xu2009automatic, lin2015complete, sun2017learning}. Most previous works rely on the hierarchical representation of simple strokes~\cite{xu2005automatic,xu2009automatic} as the basis to represent Chinese characters. For example, StrokeBank~\cite{zong2014strokebank} decomposes Chinese characters into a tree of components. FlexiFont~\cite{pan2014flexifont} scans and processes the camera-captured handwritten character images, and then formats these characters as a personalized TrueType font file. Automatic shape morphing~\cite{lian2012automatic} first generates the shape template for every character, then decomposes two given Chinese characters into strokes to establish an accurate correspondence between strokes to achieve non-rigid point set registration. More recently, awesome typography~\cite{yang2016awesome} explores the problem of generating special-effects for the typography, and exploits the statistics on the high regularity of the spatial distribution for text effects to guide the synthesis process. Zi2zi~\cite{zi2zi} considers each Chinese character as a whole and learns to transform between fonts with paired training data.
However, in the task of generating personalized Chinese handwriting font, it is difficult to obtain a large set of paired training examples.

\subsection{Image style transfer}

Current image style transfer methods can be divided into two categories, namely
descriptive neural methods based on image iteration and generative neural methods based on model iteration~\cite{jing2017neural}. Descriptive neural methods transfer the style by directly computing the gradient with respect to the source image and updating pixels in the image iteratively, while generative neural methods first optimize a generative model and produces the styled image through a single forward pass.

Neural style~\cite{gatys2015neural} is one of the most widely used descriptive neural methods for reproducing the content of an image with the style of another. It formulates style transfer as an optimization problem that combines texture synthesis with content reconstruction. Patch-based loss is added on top of content and style losses in \cite{li2016combining, chen2016fast}. 

The drawback of descriptive neural methods is that the iterative updating algorithm only works for a single image, which makes it rather time-consuming if one would like to transfer the styles of many images. In contrast, generative neural methods are faster but usually generates poorer style transfer results.


Nonetheless, our problem of generating handwritten Chinese characters does not exactly fall into the neural style transfer domain. In particular, it is difficult to define the content loss between characters in different styles since the strokes can be very different in positions and angles. For example, although $x_i$ and $y_i$ in Figure~\ref{fig:pair_unpair}(a) represent the same characters, the corresponding images differ drastically. 

\subsection{Generative adversarial networks}

\textbf{GANs.} General adversarial networks (GANs) \cite{goodfellow2014generative} are powerful generative models which have achieved impressive results in many computer vision tasks such as image inpainting \cite{iizuka2017globally} and image-to-image translation \cite{zhu2017unpaired}, as well as natural language processing tasks such as speech synthesis \cite{oord2016wavenet} and cross-language learning \cite{joty2017cross}. 
GANs formulate generative modeling as a game between two competing networks: a generator network produces synthetic data given some input noise and a discriminator network distinguishes between the generator's output and true data. Formally, the game between the generator $G$ and the discriminator $D$ has the minimax objective:
\begin{equation}
\min_{G} \max_{D} \mathbb{E}_{x \sim \mathbb{P}_r}  [\log D(x)] + \mathbb{E}_{z\sim \mathbb{P}_g}  [\log(1-D(G(z))],
\end{equation}
where $x\sim \mathbb{P}_r$ are samples from the input data and $z \sim \mathbb{P}_g$ are the random noise samples, $G(z)$ are the generated images using the neural network generator $G$, and $D(\cdot)$ gives the probability of an input being real. 

\textbf{cGANs and pix2pix}. Unlike GANs which learn a mapping from a random noise vector to an output image, conditional GANs (cGANs) learn a mapping from a random noise vector to an output image conditioning on additional information. cGANs are capable of image-to-image translation since they can condition on an input image and generate a corresponding output image. Pix2pix\cite{isola2016image} is a generic image-to-image translation algorithm using cGANs. It can produce reasonable results on a wide variety of problems. Given a training set which contains pairs of related images, pix2pix learns how to convert an image of one type into an image of another type, or vice versa.

\textbf{Zi2zi.} Zi2zi~\cite{zi2zi} uses GAN to transform Chinese characters between fonts in an end-to-end fashion, assuming no stroke label or any other auxiliary information which is usually difficult to obtain.
The network structure of zi2zi is based on pix2pix with the addition of category embedding for multiple fonts. This enables zi2zi to transform characters into several different fonts with one trained model. Zi2zi uses paired Chinese characters of the source font and the target font as the training data. However, since it is impractical to obtain a large set of paired training examples for personalized handwritten Chinese character generation, zi2zi is not applicable to our problem.



\textbf{CycleGAN.} Cycle-consistent GANs (CycleGANs) learn the image translation without paired examples~\cite{zhu2017unpaired}. Instead, it trains two generative models cycle-wise between the input and output images. In addition to the adversarial losses, cycle consistency loss is used to prevent the two generative models from contradicting each other. The default generator architecture of CycleGAN is ResNet~\cite{he2016deep}, while the default discriminator architecture is a PatchGAN classifier~\cite{isola2016image}.

\section{Our method}
\begin{figure*}[!ht]
    \begin{center}
    \includegraphics[width = \linewidth]{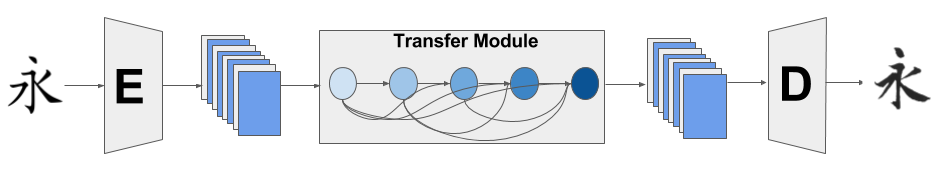}
    \end{center}
    \caption{\textbf{The architecture of the generator in DenseNet-5 CycleGAN}. The information in the image in source style is first encoded to a lower dimensional space via the encoder $E$. Then the extracted features go through a transfer module, the outputs of which can be considered as the extracted features in the target style. Finally, the output of the transfer module is decoded via the decoder $D$.}
    \label{fig:generator_architecture}
\end{figure*}

Our task is to generate handwritten Chinese characters using unpaired source and target fonts. It can be formulated as learning a mapping $G$ from the source font $X$ to the target font $Y$ given training samples $\{x_i\}_{i=1}^N$ where $x_i \in X$ and $\{y_j\}_{j=1}^M$ where $y_j \in Y$. An example of the unpaired training examples can be found in Figure~\ref{fig:pair_unpair}(b). Our objective is consistent with that of CycleGAN~\cite{zhu2017unpaired}, which is to learn image/font transformations across domains without paired training examples.

The mapping $G: X \to Y$ can be seen as the generators in the GANs except that the input of the generator is an image from the source domain rather than a random noise. An example of the architecture of generator $G$ is shown in Figure~\ref{fig:generator_architecture}. The mapping for learning the styles works as follows: the information in the image from domain $X$ is first encoded to a lower dimension via the encoder given in Table~\ref{tab: architecture}. Then the extracted features from the source domain go through a transfer module. The outputs of the transfer module can be considered as the extracted features from the target domain. Finally, the output of the transfer module is decoded via the decoder as shown in Table~\ref{tab: architecture}. 

\begin{table}[htbp]
    \begin{center}
    \small
    \begin{tabular}{c|c}
     \hline
     {\bf Module} & {\bf Specifications} \\ \hline  
     \multirow{3}{*}{Encoder} & 7$\times$ 7 Conv-Norm-ReLU, 64 filters, stride 1 \\ 
     \cline{2-2}
     & 3$\times$ 3 Conv-Norm-ReLU, 128 filters, stride 2 \\ \cline{2-2}
     & 3$\times$ 3 Conv-Norm-ReLU, 256 filters, stride 2 \\ \hline
     \multirow{5}{*}{Transfer} & Dense block, growth rate 256 \\ \cline{2-2}
     & Dense block, growth rate 256 \\ \cline{2-2}
     & Dense block, growth rate 256 \\ \cline{2-2}
     & Dense block, growth rate 256 \\ \cline{2-2}
     & Dense block, growth rate 256 \\ \hline
    \multirow{3}{*}{Decoder} & 3$\times$ 3 Deconv-Norm-ReLU, 128 filters, stride $1/2$\\ \cline{2-2}
     & 3$\times$ 3 Deconv-Norm-ReLU, 64 filters, stride $1/2$ \\ \cline{2-2}
     & 7$\times$ 7 Deconv-Norm-ReLU, 3 filters, stride 1 \\ \hline
    \end{tabular}
    \end{center}
    \caption{\textbf{The architecture and layer specifications of the encoder, transfer, and decoder modules of the generator in DenseNet CycleGAN.} Conv-Norm-ReLU represents a Convolution-InstanceNorm-ReLU layer. Deconv-Norm-ReLU represents a Fractional-strided-convolution-InstanceNorm-ReLU layer.}
    \label{tab: architecture}
\end{table} 

An adversarial discriminator $D_G$, which is a 70$\times$70 PatchGAN~\cite{isola2016image}, is used to assess the qualify of the generated images in the target domain. The second mapping $F: Y \to X$ and the corresponding discriminator $D_F$ can be defined similarly. The objective of the discriminator is to distinguish between true images from the target domain and fake images produced by the generator based on images from the source domain.

The loss function of CycleGAN contains two parts: adversarial losses and cycle consistency losses. On one hand, the adversarial losses aim to match the distribution of generated images to the data distribution in the target domain. For $G$ and $D_G$, the objective is
\begin{align}
\mathcal{L}_{\mathrm{GAN}, G}(G, D_G) &=  \mathbb{E}_{x \sim p(x)}\left[\log \left(1-D_G\left(G(x)\right)\right)\right] \nonumber\\
&+ \mathbb{E}_{y \sim p(y)}\left[\log D_G(y)\right].
\end{align}
The adversarial loss for $F$ and $D_F$ can be defined similarly.
On the other hand, the cycle consistency losses ensure that the cyclic transformation is able to bring the image back to the original state. This is defined by
\begin{align}
\mathcal{L}_{\mathrm{cycle}}(G, F) &=  \mathbb{E}_{x \sim p(x)}[||F(G(x)) - x||_1] \nonumber\\
&+  \mathbb{E}_{y \sim p(y)}[||G(F(y)) - y||_1].
\end{align}
The cycle consistency loss can be seen as a regularization and the strength of the regularization is controlled by $\lambda$. The full objective of CycleGAN is the summation of the adversarial losses for both mappings and the cycle consistency losses:
\begin{align}
    \mathcal{L}_{\mathrm{total}}(G, D_G, F, D_F) &= \mathcal{L}_{\mathrm{GAN}, G}(G, D_G) \nonumber\\
    &+ \mathcal{L}_{\mathrm{GAN}, F}(F, D_F) \nonumber\\
    &+ \lambda \mathcal{L}_{\mathrm{cycle}}(G, F).
\end{align}

The choice of the architecture of the transfer module is flexible. In the original CycleGAN~\cite{zhu2017unpaired}, the transfer module contains several Residual Network (ResNet) blocks~\cite{he2016deep}:
\begin{equation}
\mathbf{x}_{\ell}
=
H_{\ell}(\mathbf{x}_{\ell-1})
+
\mathbf{x}_{\ell-1},
\end{equation}
where $\mathbf{x}_{\ell-1}$ and $\mathbf{x}_{\ell}$ are the input and output of the $\ell$-the ResNet block, and $H_{\ell}$ represents a composite function of operations such as batch normalization (BN)~\cite{ioffe2015batch}, rectified linear units (ReLU)~\cite{glorot2011deep} and convolution. The purpose of using an identity skip-connection to bypass the non-linear transformations is to facilitate gradient back-propagation.

Apart from the ResNet, the Dense Convolutional Network (DenseNet)~\cite{huang2016densely} is among the most recent developments in convolutional neural networks.
It further improves the information flow across blocks by connecting all the blocks directly with each other.
The $\ell$-th block receives the feature maps of all preceding blocks, $\mathbf{x}_0, \ldots, \mathbf{x}_{\ell-1}$, as input,
\begin{equation}
\mathbf{x}_{\ell}
=
H_{\ell}([\mathbf{x}_0, \mathbf{x}_1, \ldots, \mathbf{x}_{\ell-1}]),
\end{equation}
where $[\mathbf{x}_0, \mathbf{x}_1, \ldots, \mathbf{x}_{\ell-1}]$ refers to the concatenation of the
feature maps produced in blocks $0, \ldots, \ell-1$. 
DenseNets achieve state-of-the-art classification accuracy across several highly competitive datasets, using fewer parameters and less computation than ResNets.

Inspired by the design of DenseNets, we incorporate DenseNet blocks in the transfer module and propose a CycleGAN with DenseNet generator architecture (DenseNet CycleGAN) to generate handwritten Chinese characters. As shown in Table~\ref{tab: architecture}, the transfer module in our DenseNet CycleGAN consists of DenseNet blocks instead of ResNet blocks.

\section{Experiments}
In this section, we evaluate our proposed method on two publicly available datasets. Furthermore, we propose content accuracy and style discrepancy as complementary evaluation metrics in addition to visual appearance. Main results are shown in this section. 

\subsection{Datasets}

\textbf{CASIA-HWDB dataset.}
The Chinese handwriting database, CASIA-HWDB~\cite{liu2011casia} is a widely used database for Chinese handwritten character recognition~\cite{zhong2015high,zhang2017online}. This database contains samples of isolated characters and handwritten texts that were produced by 1020 writers using Anoto pen on papers.

In this study, we use the HWDB1.1 dataset from the CASIA-HWDB. It contains 300 files (240 in HWDB1.1 training set and 60 in HWDB1.1 test set). Each file contains about 3000 isolated gray-scale Chinese character images written by one writer, as well as their corresponding labels. The isolated character images are resized to $128\times 128$ pixels. Other than resizing, no other data preprocessing method is performed.

\begin{figure}[ht]
    \begin{center}
    \begin{subfigure}[c]{0.19\linewidth}
    \includegraphics[width=\linewidth]{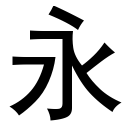}
    \caption{SIMHEI}
    \end{subfigure}
    \begin{subfigure}[c]{0.19\linewidth}
    \includegraphics[width=\linewidth]{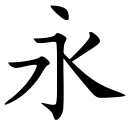}
    \caption{SIMKAI}
    \end{subfigure}
    \begin{subfigure}[c]{0.19\linewidth}
    \includegraphics[width=\linewidth]{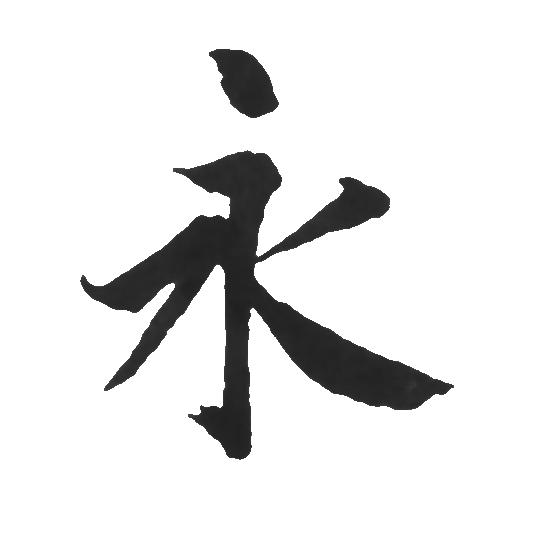}
    \caption{Lanting}
    \end{subfigure}
    \begin{subfigure}[c]{0.19\linewidth}
    \includegraphics[width=\linewidth]{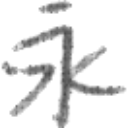}
    \caption{HW252}
    \end{subfigure}
    \begin{subfigure}[c]{0.19\linewidth}
    \includegraphics[width=\linewidth]{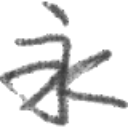}
    \caption{HW292}
    \end{subfigure}
    \end{center}
    \caption{\textbf{The character ``yong" in 5 different fonts.} (a) SIMHEI; (b) SIMKAI; (c) character in Lanting calligraphy dataset; (d) handwritten character from HW252 ({\tt 1252-c.gnt}) in HWDB1.1; (e) handwritten character from HW292 ({\tt 1292-c.gnt}) in HWDB1.1.}
    \label{fig:yong}
\end{figure}

For the task of generating handwritten characters, we use the file HW252 ({\tt 1252-c.gnt}) in the HWDB1.1 dataset as the target style, and SIMHEI font as the source style. SIMHEI is a commonly used Chinese font. Figure~\ref{fig:yong} shows the Chinese character ``yong" in 5 different styles. The first two are printed fonts, and the last three are handwritten. 

\textbf{Lanting calligraphy dataset.}
Chinese calligraphy is a form of aesthetically pleasing writing, which has been widely practiced in China and has been generally highly esteemed in the Chinese cultural sphere. In this work, we use the calligraphic work by Wang Xizhi as an example, who is generally regarded as the greatest Chinese calligrapher in history. Wang's most famous work is the \textit{Lantingji Xu}, which consists of 324 characters written in the \textit{semi-cursive style}. 

Each character in the \textit{Lantingji Xu} is first scanned and isolated from the manuscript. They are further binarized and denoised using median filtering. Finally, the characters are padded to square and resized to $128\times128$ pixels. The resulting dataset is referred to as the \textbf{Lanting calligraphy dataset}. The characters in the first column in Figure~\ref{fig:xizhi_calligraphy} are examples from the Lanting calligraphy dataset. The source font used for this task is the SIMKAI font, which is in the \textit{regular style}. The dataset can be found here: \textit{https://github.com/changebo/HCCG-CycleGAN/blob/master/lanting.zip}.

\subsection{Performance metrics}
Generative models usually lack objective evaluation criteria, which makes it difficult to quantitatively assess the quality of the images generated. To measure the performance of our handwritten character generation method on the CASIA-HWDB dataset, we propose two complementary evaluation metrics: the content accuracy and the style discrepancy. Both evaluations are based on a pre-trained network: the HCCR-GoogLeNet~\cite{zhong2015high}, which is a handwritten Chinese character classification model based on GoogLeNet~\cite{szegedy2015going}.

\textbf{Content accuracy.} 
The HCCR-GoogLeNet model is trained using the CASIA-HWDB handwritten character database with $1,020$ writers in total, including HW252. It achieved the state-of-the-art accuracy of Chinese character classification.
Inspired by the idea of the Inception score~\cite{salimans2016improved}, the pre-trained HCCR-GoogLeNet model can be used to evaluate the quality of the generated handwritten characters.
The intuition is that if the generated characters are realistic, the pre-trained HCCR-GoogLeNet will also be able to classify the generated characters correctly. In our case, characters in the target style are generated from available images in the source style. Therefore, the true labels of the generated characters are known. If the characters generated can be accurately classified by pre-trained character recognition models, to some extent it indicates that the generative model is of high quality.

Table~\ref{tab: HCCR_baseline} shows the test accuracy of HCCR-GoogLeNet on the HW252 handwritten characters and the SIMHEI font characters as a baseline. The high recognition accuracy indicates that the HCCR-GoogLeNet accuracy is a reliable metric for quality measurement. However, the equally good performance on HW252 and SIMHEI implies that this metric measures the generation quality from a single perspective: it only assesses the content quality, not the style quality. Therefore, the recognition accuracy of HCCR-GoogLeNet on the generated handwritten characters is referred to as the content accuracy.


\begin{table}[b]
\begin{center}
\begin{tabular}{ccc}
\hline
       & \textbf{Top-1 accuracy} & \textbf{Top-5 accuracy} \\ \hline
\textbf{HW252}   & 98.03\%     &      99.84\% \\ \hline
\textbf{SIMHEI} & 94.22\%     &      99.60\% \\ \hline
\end{tabular}
\end{center}
\caption{The top-1 and top-5 test accuracy of HCCR-GoogLeNet on the HW252 handwritten characters and the SIMHEI font characters. The equally good performance on HW252 and SIMHEI implies that this metric only measures the content quality.}
\label{tab: HCCR_baseline}
\end{table}

\textbf{Style discrepancy.}
To measure the discrepancy in style between the true characters in the target domain and the generated characters, we borrow the style loss in neural style transfer algorithm~\cite{gatys2015neural}.
The idea is to use the correlations between different filter activations at one layer as a style representation.
The feature correlations are given by the Gram matrix $G^{\ell} \in \mathbb{R}^{N_{\ell} \times N_{\ell}}$, where $N^{\ell}$ is the number of filters in the $\ell$-th layer, and $G_{ij}^{\ell}$ is the inner product between the vectorized feature map $i$ and $j$ in layer $\ell$:
\begin{equation}
G_{ij}^{\ell} = \sum_k F_{ik}^{\ell} F_{jk}^{\ell}.
\end{equation}
The style discrepancy is thus defined as the root-mean-square difference between the style representations of the target characters and the generated characters. Lower discrepancy corresponds to better style quality.
In our experiments, the input of Inception module $3$ in HCCR-GoogLeNet is used as layer $\ell$ to calculate the style discrepancy.

We run two baseline experiments to get an approximate of the range of the style discrepancy. 
\textbf{(a)} The style discrepancy between two randomly and equally split subsets of the HW252 handwritten dataset. Since these two subsets are written by the same person and have the same style, the result represents the lower bound of the style discrepancy. \textit{The style discrepancy lower bound is 503.77.}
\textbf{(b)} The style discrepancy between HW252 and SIMHEI. This is the style difference between the source style and the target style. It thus represents the most possible disagreement in style, and it measures the style quality of a trivial identity style transfer model. Therefore it can be regarded as an upper bound of the style discrepancy. \textit{The style discrepancy upper bound is 3006.03.}

\subsection{Implementation details}
In the experiments, we consider two types of transfer modules: ResNet with 6 blocks (ResNet-6) and DenseNet with 5 blocks (DenseNet-5). The DenseNet-5 transfer module has roughly the same number of parameters as the ResNet-6 transfer module. 

The only preprocessing procedure we used is to resize the training images to $128\times 128$ pixels; no other preprocessing methods (e.g. crop and flip) are used. For all the experiments, the regularization strength is set to $\lambda = 10$, and the Adam optimizer~\cite{kingma2014adam} with a batch size of 1 is used. The learning rate is set to $0.0002$ for the first 100 epochs and then linearly decays to $0$ over the next $100$ epochs. The number of iterations in each epoch in the experiments is the larger number of the training examples in the two styles. 


\subsection{Handwritten characters results}
We use SIMHEI font as the source style and handwritten characters in HW252 as the target style. SIMHEI and HW252 are both split to unpaired training and validation sets.
In real applications, we would like the users to only write a few Chinese characters, based on which the remaining handwritten characters can be generated with his/her personal style using our proposed method.
Therefore, the goal of this experiment is to use small training sets to train a CycleGAN model, and perform style transfer on the validation sets. 
Let $r_A$ be the split ratio of HW252, that is, $r_A$ of the characters in HW252 are randomly assigned to the training set, and the remaining characters are in the validation set. 
Similarly, $r_B$ denotes the split ratio of SIMHEI.
Both $r_A$ and $r_B$ take values in $\{5\%, 10\%, 15\%, 30\%\}$, which gives 16 combinations.
Table~\ref{tab: ResNet_quant} and \ref{tab: densenet_quant} show the top-5 content accuracy and style discrepancy for ResNet-6 and DenseNet-5, respectively.
The results indicate that both content and style quality improves with the number of increasing training data. 
In particular, when $r_A$ and $r_B$ are greater than $10\%$, the content accuracy is always greater than $80\%$. The performance of ResNet-6 and DenseNet-5 are comparable. Furthermore, since the style discrepancy ranges between 503.77 and 3006.03, the style discrepancies are on the low end of the spectrum. 

\begin{figure}[ht]
\begin{center}
  \begin{subfigure}[c]{0.24\linewidth}
    {\includegraphics[width=\linewidth]{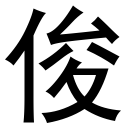}}
  \end{subfigure}
  \begin{subfigure}[c]{0.24\linewidth}
  {\includegraphics[width=\linewidth]{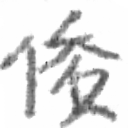}}
  \end{subfigure}
  \begin{subfigure}[c]{0.24\linewidth}
    {\includegraphics[width=\linewidth]{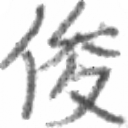}}
  \end{subfigure}
  \begin{subfigure}[c]{0.24\linewidth}
    {\includegraphics[width=\linewidth]{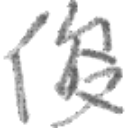}}
  \end{subfigure}
  \\
  \begin{subfigure}[c]{0.24\linewidth}
    {\includegraphics[width=\linewidth]{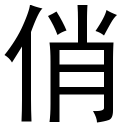}}
    \caption{SIMHEI}
  \end{subfigure}
  \begin{subfigure}[c]{0.24\linewidth}
  {\includegraphics[width=\linewidth]{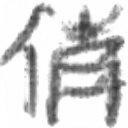}}
    \caption{DenseNet-5}
  \end{subfigure}
  \begin{subfigure}[c]{0.24\linewidth}
    {\includegraphics[width=\linewidth]{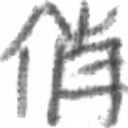}}
      \caption{ResNet-6}
\end{subfigure}
  \begin{subfigure}[c]{0.24\linewidth}
    {\includegraphics[width=\linewidth]{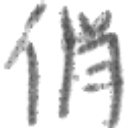}}
      \caption{HW252}
\end{subfigure}
\end{center}
\caption{\textbf{The handwritten Chinese characters}. (a) The source characters in SIMHEI font; (b) the generated characters in HW252's style using ResNet-6; (c) the generated characters in HW252's style using DenseNet-5; (d) the ground truth characters in HW252's style.} 
 \label{fig:junqiao_handwritten}  
\end{figure}

\begin{figure}
\[
\begin{array}{c|c| c}
\begin{array}{c}
\includegraphics[width=0.07\linewidth]{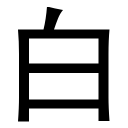}\\
\includegraphics[width=0.07\linewidth]{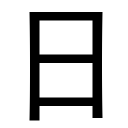}\\
\includegraphics[width=0.07\linewidth]{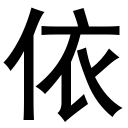}\\
\includegraphics[width=0.07\linewidth]{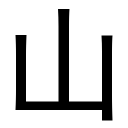}\\
\includegraphics[width=0.07\linewidth]{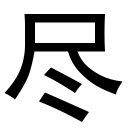}\\
\includegraphics[width=0.07\linewidth]{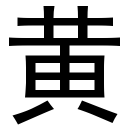}\\
\includegraphics[width=0.07\linewidth]{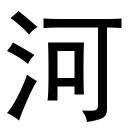}\\
\includegraphics[width=0.07\linewidth]{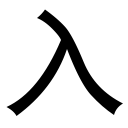}\\
\includegraphics[width=0.07\linewidth]{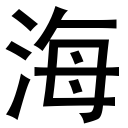}\\
\includegraphics[width=0.07\linewidth]{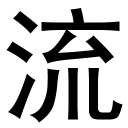}\\
\includegraphics[width=0.07\linewidth]{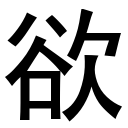}\\
\includegraphics[width=0.07\linewidth]{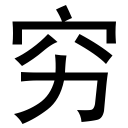}\\
\includegraphics[width=0.07\linewidth]{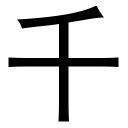}\\
\includegraphics[width=0.07\linewidth]{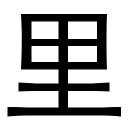}\\
\includegraphics[width=0.07\linewidth]{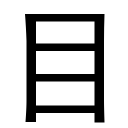}\\
\includegraphics[width=0.07\linewidth]{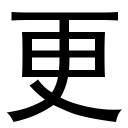}\\
\includegraphics[width=0.07\linewidth]{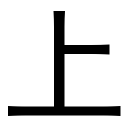}\\
\includegraphics[width=0.07\linewidth]{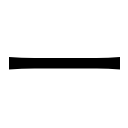}\\
\includegraphics[width=0.07\linewidth]{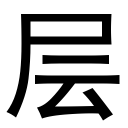}\\
\includegraphics[width=0.07\linewidth]{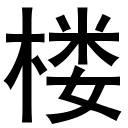}\\
\\
\end{array}&
\begin{array}{cc}
\includegraphics[width=0.07\linewidth]{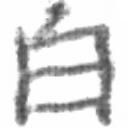}&\includegraphics[width=0.07\linewidth]{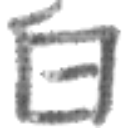}\\
\includegraphics[width=0.07\linewidth]{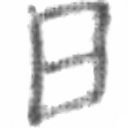}&\includegraphics[width=0.07\linewidth]{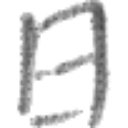}\\
\includegraphics[width=0.07\linewidth]{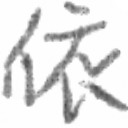}&\includegraphics[width=0.07\linewidth]{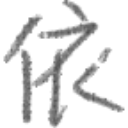}\\
\includegraphics[width=0.07\linewidth]{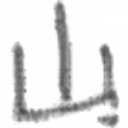}&\includegraphics[width=0.07\linewidth]{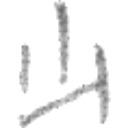}\\
\includegraphics[width=0.07\linewidth]{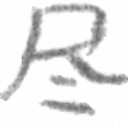}&\includegraphics[width=0.07\linewidth]{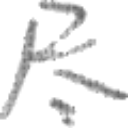}\\
\includegraphics[width=0.07\linewidth]{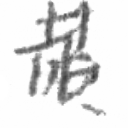}&\includegraphics[width=0.07\linewidth]{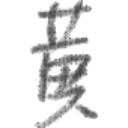}\\
\includegraphics[width=0.07\linewidth]{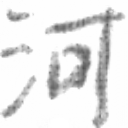}&\includegraphics[width=0.07\linewidth]{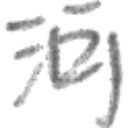}\\
\includegraphics[width=0.07\linewidth]{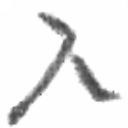}&\includegraphics[width=0.07\linewidth]{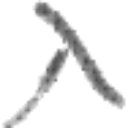}\\
\includegraphics[width=0.07\linewidth]{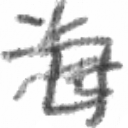}&\includegraphics[width=0.07\linewidth]{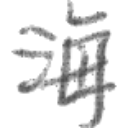}\\
\includegraphics[width=0.07\linewidth]{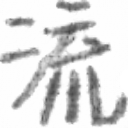}&\includegraphics[width=0.07\linewidth]{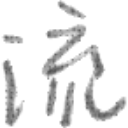}\\
\includegraphics[width=0.07\linewidth]{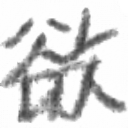}&\includegraphics[width=0.07\linewidth]{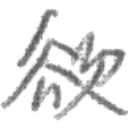}\\
\includegraphics[width=0.07\linewidth]{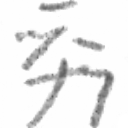}&\includegraphics[width=0.07\linewidth]{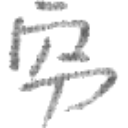}\\
\includegraphics[width=0.07\linewidth]{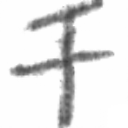}&\includegraphics[width=0.07\linewidth]{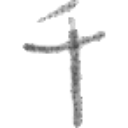}\\
\includegraphics[width=0.07\linewidth]{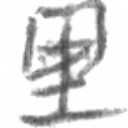}&\includegraphics[width=0.07\linewidth]{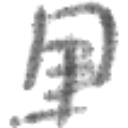}\\
\includegraphics[width=0.07\linewidth]{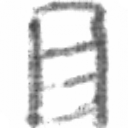}&\includegraphics[width=0.07\linewidth]{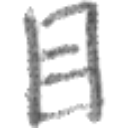}\\
\includegraphics[width=0.07\linewidth]{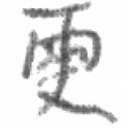}&\includegraphics[width=0.07\linewidth]{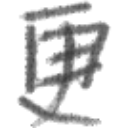}\\
\includegraphics[width=0.07\linewidth]{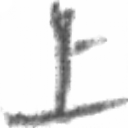}&\includegraphics[width=0.07\linewidth]{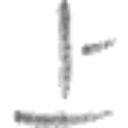}\\
\includegraphics[width=0.07\linewidth]{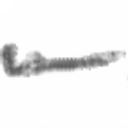}&\includegraphics[width=0.07\linewidth]{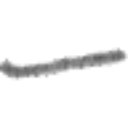}\\
\includegraphics[width=0.07\linewidth]{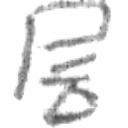}&\includegraphics[width=0.07\linewidth]{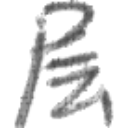}\\
\includegraphics[width=0.07\linewidth]{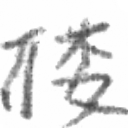}&\includegraphics[width=0.07\linewidth]{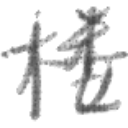}\\
\text{Generated} & \text{Truth}\\
\end{array}&
\begin{array}{cc}
\includegraphics[width=0.07\linewidth]{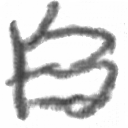}&\includegraphics[width=0.07\linewidth]{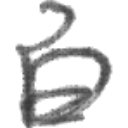}\\
\includegraphics[width=0.07\linewidth]{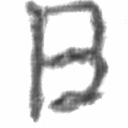}&\includegraphics[width=0.07\linewidth]{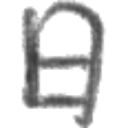}\\
\includegraphics[width=0.07\linewidth]{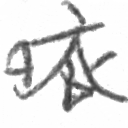}&\includegraphics[width=0.07\linewidth]{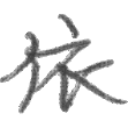}\\
\includegraphics[width=0.07\linewidth]{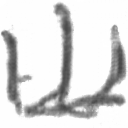}&\includegraphics[width=0.07\linewidth]{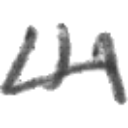}\\
\includegraphics[width=0.07\linewidth]{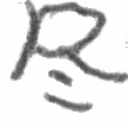}&\includegraphics[width=0.07\linewidth]{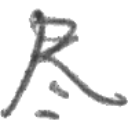}\\
\includegraphics[width=0.07\linewidth]{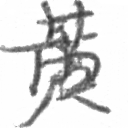}&\includegraphics[width=0.07\linewidth]{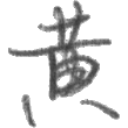}\\
\includegraphics[width=0.07\linewidth]{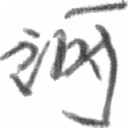}&\includegraphics[width=0.07\linewidth]{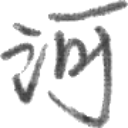}\\
\includegraphics[width=0.07\linewidth]{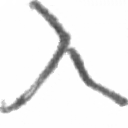}&\includegraphics[width=0.07\linewidth]{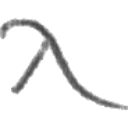}\\
\includegraphics[width=0.07\linewidth]{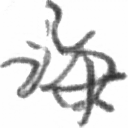}&\includegraphics[width=0.07\linewidth]{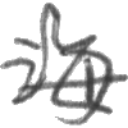}\\
\includegraphics[width=0.07\linewidth]{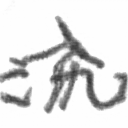}&\includegraphics[width=0.07\linewidth]{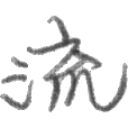}\\
\includegraphics[width=0.07\linewidth]{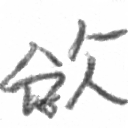}&\includegraphics[width=0.07\linewidth]{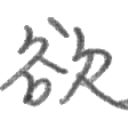}\\
\includegraphics[width=0.07\linewidth]{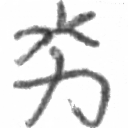}&\includegraphics[width=0.07\linewidth]{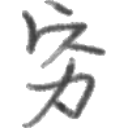}\\
\includegraphics[width=0.07\linewidth]{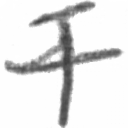}&\includegraphics[width=0.07\linewidth]{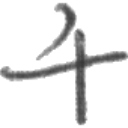}\\
\includegraphics[width=0.07\linewidth]{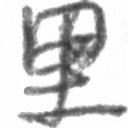}&\includegraphics[width=0.07\linewidth]{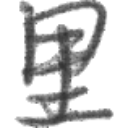}\\
\includegraphics[width=0.07\linewidth]{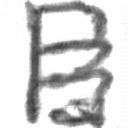}&\includegraphics[width=0.07\linewidth]{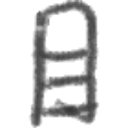}\\
\includegraphics[width=0.07\linewidth]{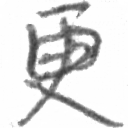}&\includegraphics[width=0.07\linewidth]{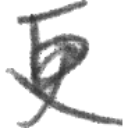}\\
\includegraphics[width=0.07\linewidth]{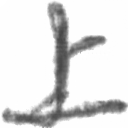}&\includegraphics[width=0.07\linewidth]{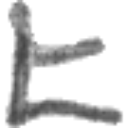}\\
\includegraphics[width=0.07\linewidth]{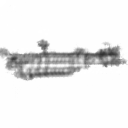}&\includegraphics[width=0.07\linewidth]{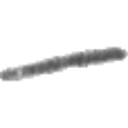}\\
\includegraphics[width=0.07\linewidth]{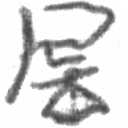}&\includegraphics[width=0.07\linewidth]{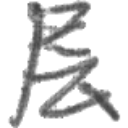}\\
\includegraphics[width=0.07\linewidth]{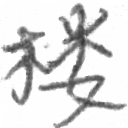}&\includegraphics[width=0.07\linewidth]{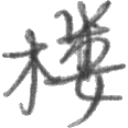}\\
\text{Generated} & \text{Truth}\\
\end{array}\\
\text{(a)}& \text{(b) HW252}& \text{(c) HW292} \\
\end{array}
\]
\caption{\textbf{A famous Chinese poem entitled ``On the Stork Tower'' generated by our proposed method.}
(a) SIMHEI is the source style; (b) and (c) are two handwritten styles. The generated characters are clearly recognizable with personalized style.}
\label{fig: dengguanquelou}
\end{figure}

Figure~\ref{fig:junqiao_handwritten} shows the generated handwritten characters using DenseNet-5 and ResNet-6, as well as the source and target styles. 
The background of the generated characters is clear and the contents are perfectly recognizable. The style of the strokes is noticeably different from that of the source font. The composition of the radicals and the blurry boundaries of the strokes highly resemble the writer's handwriting style.
Figure~\ref{fig: dengguanquelou} shows a famous Chinese poem ``On the Stork Tower'' composed of characters generated by ResNet-6 CycleGAN. All the characters are clearly recognizable with personalized style.

\begin{table}[hbp]
\begin{center}
\begin{tabular}{c|c|c|c|c|c}
\hline
\multicolumn{2}{l|}{\multirow{2}{*}{\begin{tabular}[c]{@{}l@{}}ResNet-6\end{tabular}}} & \multicolumn{4}{c}{$r_A$}          \\ 
\cline{3-6} 
\multicolumn{2}{l|}{} & 5\%    & 10\%     & 15\%    & 30\%     \\ 
\hline
\multirow{8}{*}{$r_B$}& \multirow{2}{*}{5\%} & 83.85\% & 72.83\% & 56.32\% & 49.31\%\\
                   &                          & 742.93  & 786.83  & 970.98  & 1225.86        \\
\cline{2-6} 
                   & \multirow{2}{*}{10\%}     & 77.92\% & 87.30\% & 86.98\% & 80.88\%\\
                   &                          & 836.46  & 758.73  & 697.79  & 702.45      \\
\cline{2-6} 
                   & \multirow{2}{*}{15\%}    & 78.31\% & 81.17\% & 89.41\% & 86.59\%\\
                   &                          & 811.75  & 700.12  & 709.78  & 530.31 \\
\cline{2-6} 
                   & \multirow{2}{*}{30\%}     & 82.08\% & 86.34\% & 89.19\% & 86.49\% \\
                   &                          & 885.91  & 752.86  & 761.41  & 613.44 \\
\hline
\end{tabular}
\end{center}
\caption{\textbf{ResNet-6 results.} Top-5 content accuracy and style discrepancy on HW252 and SIMHEI font. $r_A$ and $r_B$ represent the split ratio of HW252 and SIMHEI respectively.}
\label{tab: ResNet_quant}
\end{table}

\begin{table}[htbp]
\begin{center}
\begin{tabular}{c|c|c|c|c|c}
\hline
\multicolumn{2}{l|}{\multirow{2}{*}{\begin{tabular}[c]{@{}l@{}}DenseNet-5\end{tabular}}} & \multicolumn{4}{c}{$r_A$}          \\ 
\cline{3-6} 
\multicolumn{2}{l|}{} & 5\%    & 10\%     & 15\%    & 30\%     \\ 
\hline
\multirow{8}{*}{$r_B$}& \multirow{2}{*}{5\%} & 76.67\% & 58.90\% & 60.75\% &  15.78\% \\
                   &                          & 839.22  & 1075.58  & 1326.33  &   2016.11 \\
\cline{2-6} 
                   & \multirow{2}{*}{10\%}     & 84.02\% & 83.01\% & 82.39\% &  84.46\%  \\
                   &                          & 840.00 & 761.60  & 844.01& 576.14        \\
\cline{2-6} 
                   & \multirow{2}{*}{15\%}    & 84.21\% & 82.33\% & 90.47\% &  82.14\%  \\
                   &                          & 792.23  & 725.15  & 723.82      &   700.19    \\
\cline{2-6} 
                   & \multirow{2}{*}{30\%}     & 81.43\% & 90.03\% & 89.04\% &  88.55\% \\
                   &                          & 792.89  & 703.96  & 759.20  &  594.876 \\
\hline
\end{tabular}
\end{center}
\caption{\textbf{DenseNet-5 results.} Top-5 content accuracy and style discrepancy on HW252 and SIMHEI font. $r_A$ and $r_B$ represent the split ratio of HW252 and SIMHEI respectively.}
\label{tab: densenet_quant}
\end{table}

\subsection{Calligraphy results}
\begin{figure}[ht]
\begin{center}
  \begin{subfigure}[c]{0.24\linewidth}
    {\includegraphics[width=\linewidth]{figures/yonghejiunian/true/0001.jpg}}
  \end{subfigure}
  \begin{subfigure}[c]{0.24\linewidth}
  {\includegraphics[width=\linewidth]{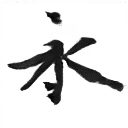}}
  \end{subfigure}
  \begin{subfigure}[c]{0.24\linewidth}
    {\includegraphics[width=\linewidth]{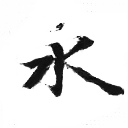}}
  \end{subfigure}
  \\
  \begin{subfigure}[c]{0.24\linewidth}
    {\includegraphics[width=\linewidth]{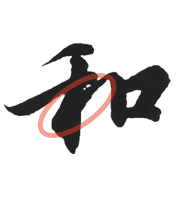}}
  \end{subfigure}
  \begin{subfigure}[c]{0.24\linewidth}
  {\includegraphics[width=\linewidth]{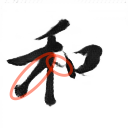}}
  \end{subfigure}
  \begin{subfigure}[c]{0.24\linewidth}
    {\includegraphics[width=\linewidth]{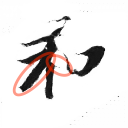}}
  \end{subfigure}
  \\
  \begin{subfigure}[c]{0.24\linewidth}
    {\includegraphics[width=\linewidth]{figures/yonghejiunian/true/0003.jpg}}
  \end{subfigure}
  \begin{subfigure}[c]{0.24\linewidth}
  {\includegraphics[width=\linewidth]{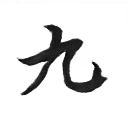}}
  \end{subfigure}
  \begin{subfigure}[c]{0.24\linewidth}
    {\includegraphics[width=\linewidth]{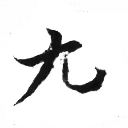}}
  \end{subfigure}
  \\
  \begin{subfigure}[c]{0.24\linewidth}
    {\includegraphics[width=\linewidth]{figures/yonghejiunian/true/0004.jpg}}
    \caption{Lanting}
  \end{subfigure}
  \begin{subfigure}[c]{0.24\linewidth}
  {\includegraphics[width=\linewidth]{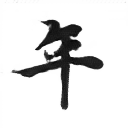}}
  \caption{DenseNet-5}
  \end{subfigure}
  \begin{subfigure}[c]{0.24\linewidth}
    {\includegraphics[width=\linewidth]{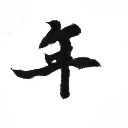}}
  \caption{ResNet-6}
  \end{subfigure}
\end{center}
\caption{\textbf{The Chinese calligraphy characters in Lanting calligraphy dataset}. (a) The ground truth characters; (b) the generated characters in Wang Xizhi's style using DenseNet-5; (c) the generated characters in Wang Xizhi's style using ResNet-6. Note that for the character ``he'' in the second row, the strokes ``throw-away'' and ``press-down'' in the red circle are simplified to a single ``break'' stroke ihe ground truth characters, but they are generated as separate strokes by DenseNet and ResNet CycleGAN.} 
 \label{fig:xizhi_calligraphy}  
\end{figure}

We use SIMKAI as the source font for the calligraphy generation in Wang Xizhi's style.
Overall, DenseNet marginally outperforms ResNet in style. 
Figure~\ref{fig:xizhi_calligraphy} shows the ground truth as well as the generated calligraphy characters of the first four characters in the Lanting calligraphy dataset given unpaired training data. It can be seen that CycleGAN with both DenseNet and ResNet generator captures the overall writing style of Wang Xizhi and generates reasonable outputs. Compared with ResNet, DenseNet tends to generate fewer cases of missing strokes (e.g., the dot in the first character) or extra strokes (e.g., the extra throw-away in the second character). Nevertheless, both DenseNet and ResNet generators fail to learn certain features of Wang's semi-cursive style. For example, for the character ``he''  in the second row, the strokes ``throw-away'' (falling leftwards) and ``press-down'' (falling rightwards) are usually simplified to a single ``break'' (a change in direction) stroke in semi-cursive script. This feature is not learned by the CycleGAN model: both DenseNet and ResNet generate separate ``throw-away'' and ``press-down'' strokes.

The last two columns in Figure~\ref{fig: wangxizhi} are also examples of the generated Chinese calligraphy in Wang Xizhi's style. These characters are not in the original Lanting calligraphy dataset and the generated characters show satisfactory performance.

\subsection{Comparison with neural style transfer}

\begin{figure}[htp]
\begin{center}
  \begin{subfigure}[c]{0.24\linewidth}
    {\includegraphics[width=\linewidth]{figures/junqiao/simhei/20426.png}}
  \end{subfigure}
  \begin{subfigure}[c]{0.24\linewidth}
  {\includegraphics[width=\linewidth]{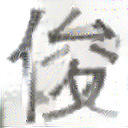}}
  \end{subfigure}
  \begin{subfigure}[c]{0.24\linewidth}
    {\includegraphics[width=\linewidth]{figures/junqiao/1252-true/20426.png}}
  \end{subfigure}
  \\
  \begin{subfigure}[c]{0.24\linewidth}
    {\includegraphics[width=\linewidth]{figures/junqiao/simhei/20431.png}}
    \caption{SIMHEI}
  \end{subfigure}
  \begin{subfigure}[c]{0.24\linewidth}
  {\includegraphics[width=\linewidth]{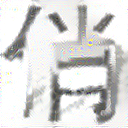}}
    \caption{Neural style}
  \end{subfigure}
  \begin{subfigure}[c]{0.24\linewidth}
    {\includegraphics[width=\linewidth]{figures/junqiao/1252-true/20431.png}}
      \caption{HW252}
\end{subfigure}
\end{center}
\caption{\textbf{The results of neural style transfer}. (a) The source characters in SIMHEI font; (b) characters generated by neural style transfer; (c) the ground truth characters in HW252's style. The style of the generated images is almost identical to the source font and the background is blurry. } 
\label{fig:junqiao_neural_style}  
\end{figure}

We compare our method with neural style transfer~\cite{gatys2015neural} by using VGG-19 as the pre-trained network. We use layer {\tt relu4\_2} of VGG-19 for content loss, and layer {\tt relu1\_1}, {\tt relu2\_1}, {\tt relu3\_1}, {\tt relu4\_1}, and {\tt relu5\_1} for style loss.
Two characters in SIMHEI font are used as the content images, and 30\% of the characters are randomly selected from the HW252 dataset as style images. This is the same as the setting of $r_A = 30\%$. The generated results by neural style transfer are shown in Figure~\ref{fig:junqiao_neural_style}. It can be seen the style of the generated images is almost identical to the source font, implying that the learned transform is trivial.
Furthermore, the background of the generated images is blurry. 
The overall visual quality is worse than those generated by our method.

\section{Conclusion}
In this work, we formulate the Chinese handwritten character generation problem as learning a mapping from an existing printed font to a personalized handwritten style. We present DenseNet CycleGAN to solve this problem, and our method uses DenseNet as part of the CycleGAN generator to improve the generation quality. The proposed method is compared with the original CycleGAN with ResNet blocks and Neural style transfer. We evaluate the proposed method on the CASIA dataset and the newly introduced Lanting calligraphy dataset. Furthermore, we propose two novel Chinese character generation performance evaluation metrics content accuracy and style discrepancy for quantitatively assessing the quality of generated Chinese handwritten character. Extensive experimental results demonstrate the efficacy of our method, showing superior or on-par performance. 

\noindent
\textbf{Acknowledgements.} The authors would like to thank NVIDIA for the hardware donation from GPU Grant Program. 

{\small
\bibliographystyle{ieee}
\bibliography{egbib}
}

\end{document}